\documentclass{article}
\usepackage{tccml_iclr2025_conference}
\usepackage{graphicx} % Required for inserting images

\usepackage{verbatim}
\usepackage{url}
\usepackage{booktabs}  % For better-looking tables
\usepackage{amsmath}   % For math symbols
\usepackage{subcaption}
\usepackage{float}    % For [H] specifier

\title{Fine Flood Forecasts: Incorporating local
data into global models through fine-tuning}
\date{January 2025}
\author{
Emil Ryd$^{1}$, Grey Nearing$^{2}$ \\
$^{1}$University of Oxford \\
$^{2}$Google Research\\
\texttt{\{emil.ryd@new.ox.ac.uk, gsnearing@google.com\}}
}

\begin{document}

\maketitle

\begin{abstract}
Floods are the most common form of natural disaster and accurate flood forecasting is essential for early warning systems. Previous work has shown that machine learning (ML) models are a promising way to improve flood predictions when trained on large, geographically-diverse datasets. This requirement of global training can result in a loss of ownership for national forecasters who cannot easily adapt the models to improve performance in their region, preventing ML models from being operationally deployed. Furthermore, traditional hydrology research with physics-based models suggests that  local data -- which in many cases is only accessible to local agencies -- is valuable for improving model performance. To address these concerns, we demonstrate a methodology of pre-training a model on a large, global dataset and then fine-tuning that model on data from individual basins. This results in performance increases, validating our hypothesis that there is extra information to be captured in local data. In particular, we show that performance increases are most significant in watersheds that underperform during global training. We provide a roadmap for national forecasters who wish to take ownership of global models using their own data, aiming to lower the barrier to operational deployment of ML-based hydrological forecast systems\footnote{Our code is available at \url{https://github.com/EmilRyd/Fine-Flood-Forecasts.git}}.

\end{abstract}
\section{Introduction}\label{intro}
Climate change is causing an increase in the intensity and frequency of heavy precipitation
\citep{Min2011}, increasing river flooding \citep{Alifu2022}. Floods are already the most frequently occurring natural disaster, having affected 2.5 billion people between 1994 and 2013 \citep{cred2015human}. The rate of floods has more than doubled since 2000 \citep{WMO2021}, and is expected to increase further due to anthropogenic climate change. 
%Alarmingly, low-income countries are most exposed to flooding disasters
%\citep{Rentschler2022}. 
The World Bank estimates that improving the flood early warning systems of developing countries to the standards of developed countries would save 23,000 lives per year on average \citep{Hallegatte2012}. In this paper, we focus specifically on the task of river-based flood prediction, and more specifically, on models that provide short-perm predictions of river flow volume.

Machine learning (ML) models trained on large global datasets have been shown to be effective at generating accurate flood forecasts \citep{Kratzert2019, Nearing2024}. This type of model performs best when trained on data from a large and diverse set of geographical locations \citep{Kratzert2024}. However, it is difficult for local flood forecasting agencies to develop and train prediction models that require global datasets. Currently, such models are instead trained and operated by multinational organizations like Google (e.g., \url{g.co/floodhub}). This implies a potential loss of ownership for the national hydromet agencies who may have an institutional preference for using models calibrated or developed locally vs. global models provided by a third party. Furthermore, national hydromet agencies often have proprietary data for their own geographical region, which they may not wish to share publicly, meaning that a global model trained by a third party would not have been trained on data from their local watersheds. 

While training a model on as much data as possible makes sense from an ML perspective, it is contrary to intuitions derived from decades of research in the hydrological sciences, where individual basin calibration traditionally provides the best forecasts \citep{Hrachowitz2013, GloFAS2018}. Based on this intuition and the needs (discussed above) of many national and local flood forecasting agencies to use their own data and models, we propose fine-tuning a globally pre-trained model using data from individual basins. Fine-tuning is a well-known practice within the ML community, and for us serves the purpose of producing models aimed at high-quality local prediction while still capturing the advantages of large-sample training. The purpose of this paper is to highlight how this practice can help facilitate trust, reliability, and adoption of best-in-class flood forecasting tools.

What we describe and test in this paper is a relatively simple workflow facilitated by open, public ML-based hydrology models, which can be adopted by local agencies. Fine-tuning requires some amount of technical expertise, but does not require large computing resources. We provide the necessary open-source pre-trained global model as well as an outline of a fine-tuning procedure that agencies might use or adapt for their needs (see Appendix \ref{fine-tuning guide} for details).% Our contributions are:
%\begin{enumerate}
%    \item We demonstrate that large-scale fine-tuning of global flood forecast models results in statistically significant performance increases.
%    \item We show that our methodology works particularly well for locations where the globally trained model does not perform well.
%    \item We provide a step-by-step guide and code for national hydromet agencies to fine-tune globally trained, ML-based hydrology models.
%\end{enumerate}

\section{Methodology}\label{methodology}
\subsection{Data}\label{data}
\begin{figure}[ht]
    \centering
    % First subfigure
    \begin{subfigure}{0.68\linewidth}
        \centering \includegraphics[width=\linewidth] {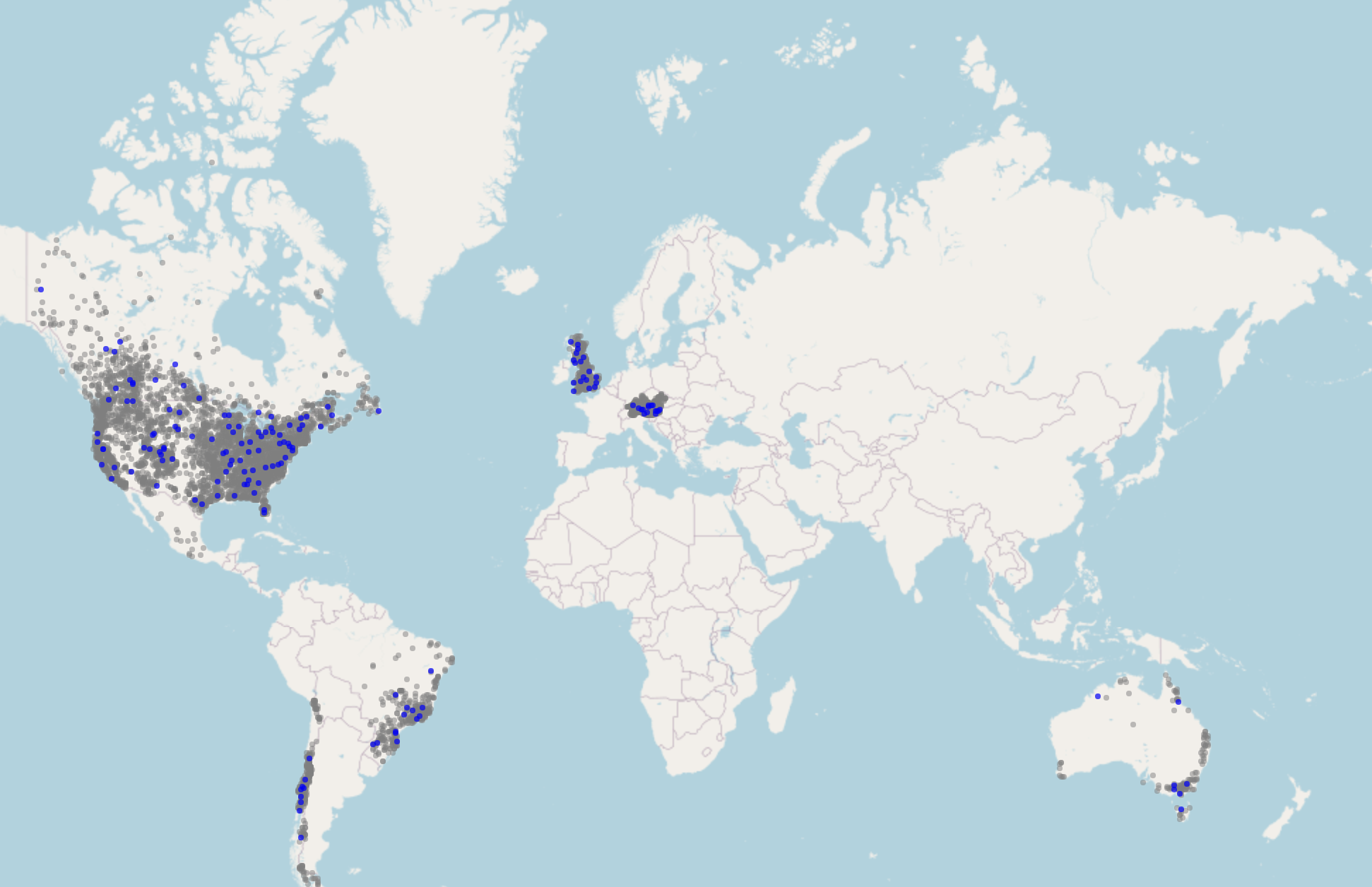} 
    \end{subfigure}
    % Second subfigure
    \begin{subfigure}{0.26\linewidth}
        \centering
\includegraphics[width=\linewidth]{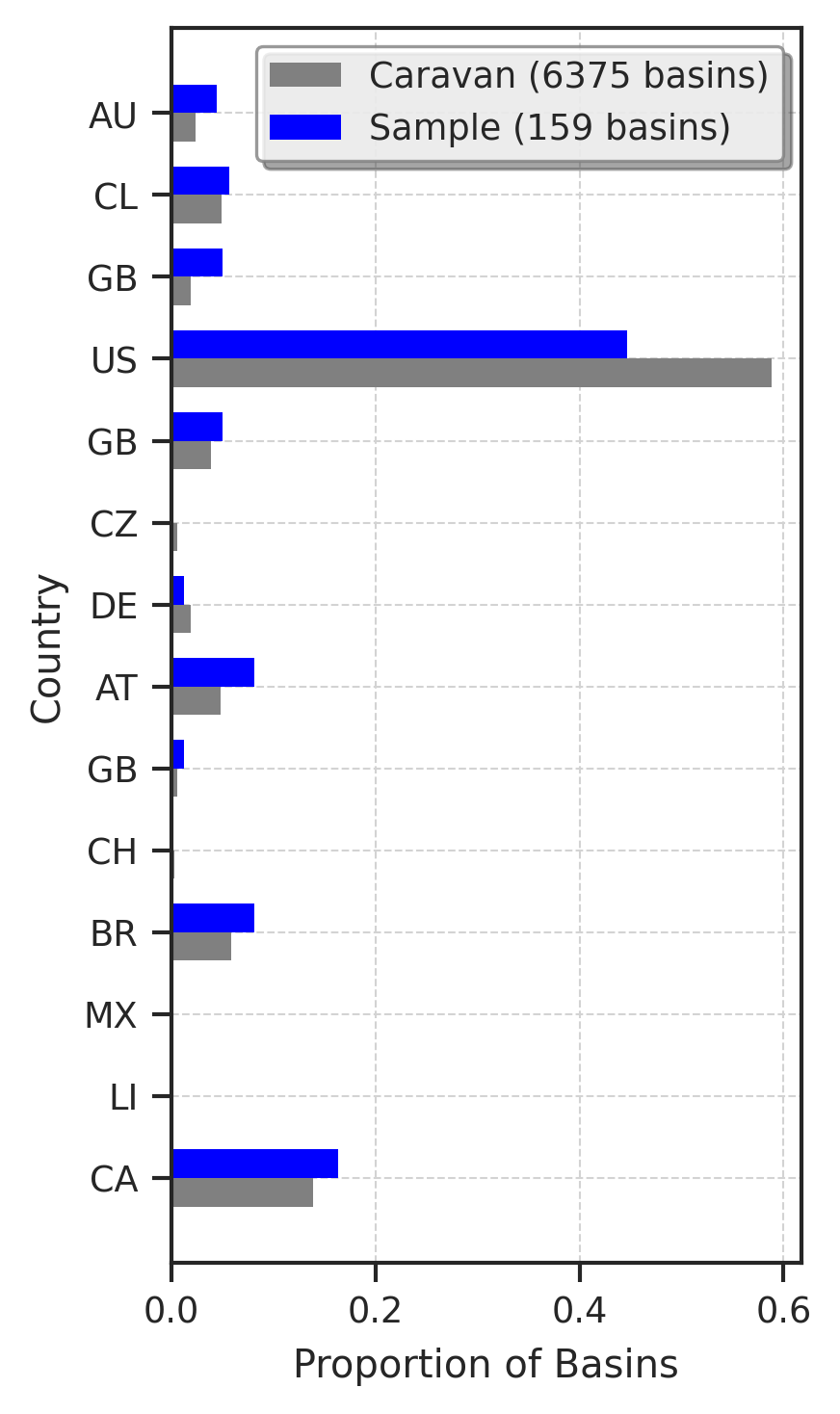}
    \end{subfigure} 
\caption{Left: world map showing all 6375 in the Caravan dataset (grey), and our 159 randomly sampled basins (blue). Right: bar chart showing how well each country is represented in our sample compared to the dataset as a whole.}

\label{fig:data_visualization}
\end{figure}

As input data, we use the global Caravan dataset \cite{kratzert2023caravan}, which contains atmospheric variables (e.g., precipitation, temperature) and catchment-specific hydrological variables (e.g. climatological soil moisture, elevation), along with globally sourced streamflow data, mostly contributed by national hydromet agencies. 
%All atmospheric input data in Caravan were collected from the global ERA-5 dataset \citep{Hersbach2020}.
%Basin drainage areas were constructed from HyBas polygons \cite{Lehner2013}. 
Currently, the Caravan dataset consists of 22,732 unique catchments \citep{Farber2024}, however to ensure continuity with previous work \citep{roi_cohen_morin_2024, kratzert2023caravan} we trained on an older, reduced version of Caravan containing $6375$ basins. Since fine-tuning on every basin in the dataset would be computationally expensive, we fine-tune on a random sample of $159$ basins, see figure \ref{fig:data_visualization}.
\vspace{-10pt}

\subsection{Model architecture \& training details}\label{model arch}
We pre-train a variant of a long short-term memory (LSTM) model, following Kratzert et al. \citep{Kratzert2018}, on all $6375$ 
\begin{comment}
    check numbers
\end{comment} 
basins. %from the global Caravan dataset \citep{kratzert2023caravan}. 
LSTMs \citep{Hochreiter1997} are a variant of recursive neural networks (RNNs) \citep{Rumelhart1986}, commonly used in temporal prediction tasks. LSTMs are suitable for streamflow prediction since they have an internal state space, which is similar to how we conceptualize physical systems such as watersheds \citep{Kratzert2018}. This allows LSTM memory cells to model hydrological dynamics such as reservoirs and storages \citep{Kratzert2019_interp, Lees2022}. Transformers \cite{Vaswani2017} have been tried, yet LSTMs remain the state-of-the-art within streamflow prediction with ML.%, hence we use them here.
The concepts outlined here are, in principle, applicable to most types of ML models.

Our LSTM model takes as input a time series of atmospheric and hydrological variables stretching $365$ days back, and then outputs a single prediction of streamflow (i.e. volume of water flowing per unit time) for the following day, same as previous work \citep{kratzert2022neuralhydrology, Kratzert2018, Nearing2024}.
We used the model configuration from Roi-Cohen and Morin \citep{roi_cohen_morin_2024}, and trained the model using the open-source Neuralhydrology library \cite{kratzert2022neuralhydrology}. Training the 260K parameter model takes 30 wall clock hours on a single NVIDIA V100 GPU.
\begin{comment}
    check if true
\end{comment}

For each basin in our sample (see section \ref{data}), we swept over $50$ sets of hyperparameters using Tree-structured Parzen estimator \citep{Watanabe2023TreestructuredPE} from the hyperopt library \citep{Bergstra2013}, selecting the hyperparameters with the best validation set performance. To ensure that our results are robust across different model instances, we ran this procedure across 8 pre-trained model instances trained with different random seeds, generating 1272 data points in total. Further information on the model configuration and training setup is given in Appendix \ref{model config}. 

\subsection{Metrics}\label{metrics}

We focus on evaluating models using the Nash-Sutcliffe efficiency (NSE) and the Kling-Gupta efficency (KGE) metrics. These are arguably the main metrics used in assessing hydrological models, and are both related to standard correlation and bias metrics \citep{Gupta2009, Knoben2019}. These metrics measure the accuracy of streamflow predictions, and a score closer to 1 is better. However, these two metrics alone are generally not sufficient for a complete hydrological analysis, and a larger suite of performance metrics is reported in Appendix \ref{more results}.

\section{Results}\label{results}
\subsection{Fine-tuning improvements}\label{Fine-tuning improvements} To evaluate the general performance improvements caused by fine-tuning, we compare the performance of the pre-trained model with its fine-tuned counterpart (see table \ref{base performance}). The results confirm our assumption (from \citep{Kratzert2024})  that training a model on a single basin gives worse performance than training a model on a larger, multi-basin dataset. Although this is counterintuitive from a hydrological standpoint (where basin-specific calibration is traditionally crucial), it is a common occurrence across almost all applications of deep learning. 
\begin{table}[H]
\caption{Model performance comparison for fine-tuned, pre-trained, and single-basin trained model. Best score in bold, second best in italics.}\label{base performance}

\label{tab:model-comparison}
\centering
\renewcommand{\arraystretch}{1.3}  % Increases vertical spacing
\small  % Reduce font size
\begin{tabular}{l@{\hspace{4pt}}c@{\hspace{4pt}}c@{\hspace{4pt}}c@{\hspace{4pt}}c@{\hspace{4pt}}c@{\hspace{4pt}}c}  % Reduced horizontal spacing
Model & \rotatebox{0}{NSE (mean)} & \rotatebox{0}{NSE (median)} & \rotatebox{0}{KGE (mean)} & \rotatebox{0}{KGE (median)} \\
\midrule
Single-basin \footnotemark[1] & 0.358 ($\pm$ 0.106) & 0.541 ($\pm$ 0.251)  & 0.331 ($\pm$ 0.197)  & 0.609 ($\pm$ 0.197)\\
Pre-trained & 0.473 ($\pm$ 0.019) & 0.583 ($\pm$ 0.016) & 0.520 ($\pm$ 0.007) & 0.683 ($\pm$ 0.010)\\
Fine-tuned & \textbf{0.541} ($\pm$ 0.017) & \textbf{0.625} ($\pm$ 0.013) & \textbf{0.599} ($\pm$ 0.042) & \textbf{0.709} ($\pm$ 0.015)\\
\end{tabular}
\end{table}
\footnotetext[1]{Only trained for 147 out of the 159 basins due to some basins not having any data within our pre-defined training set.}
\vspace{-10pt}

Fine-tuning provides a significant increase in prediction skill over the pre-trained model. We found an increase of $0.068$ $(14\%)$ and $0.079$ $(15\%)$ in the mean (over the 159 fine-tuned basins) NSE and KGE, respectively. The median increases $0.042$ $(7\%)$ for NSE and $0.026$ $(4\%)$ for KGE. This indicates that much of the improvement is happening at the tail end. The fact that the pre-trained median NSE and KGE are higher than the mean also implies that there is a significant tail end of underperforming basins,  and as we show in section \ref{correlation with pre-trained skill}, improvement from fine-tuning is particularly large in these.

To put these improvements in context, Nearing et al. \citep{Nearing2022} found an $8\%$ improvement to median NSE scores in 539 basins within the continental United States by assimilating near-real-time streamflow data. Fine-tuning uses only historical data whereas data assimilation requires near real-time data and is therefore significantly more difficult and costly to implement (real-time data is harder to access and more difficult to ingest into ML workflows). Fine-tuning therefore represents a substantial performance gain using an approach that has less demanding data requirements than data assimilation. We use data assimilation as a point of reference because, as far as we are aware, no other published method for increasing performance of LSTM-based streamflow models has shown as much skill improvement.
\vspace{-1pt}
\subsection{Correlation with pre-trained skill}\label{correlation with pre-trained skill} 
\begin{figure}[ht]
    \centering
    % First subfigure
    \begin{subfigure}{0.49\linewidth}
        \centering \includegraphics[width=\linewidth] {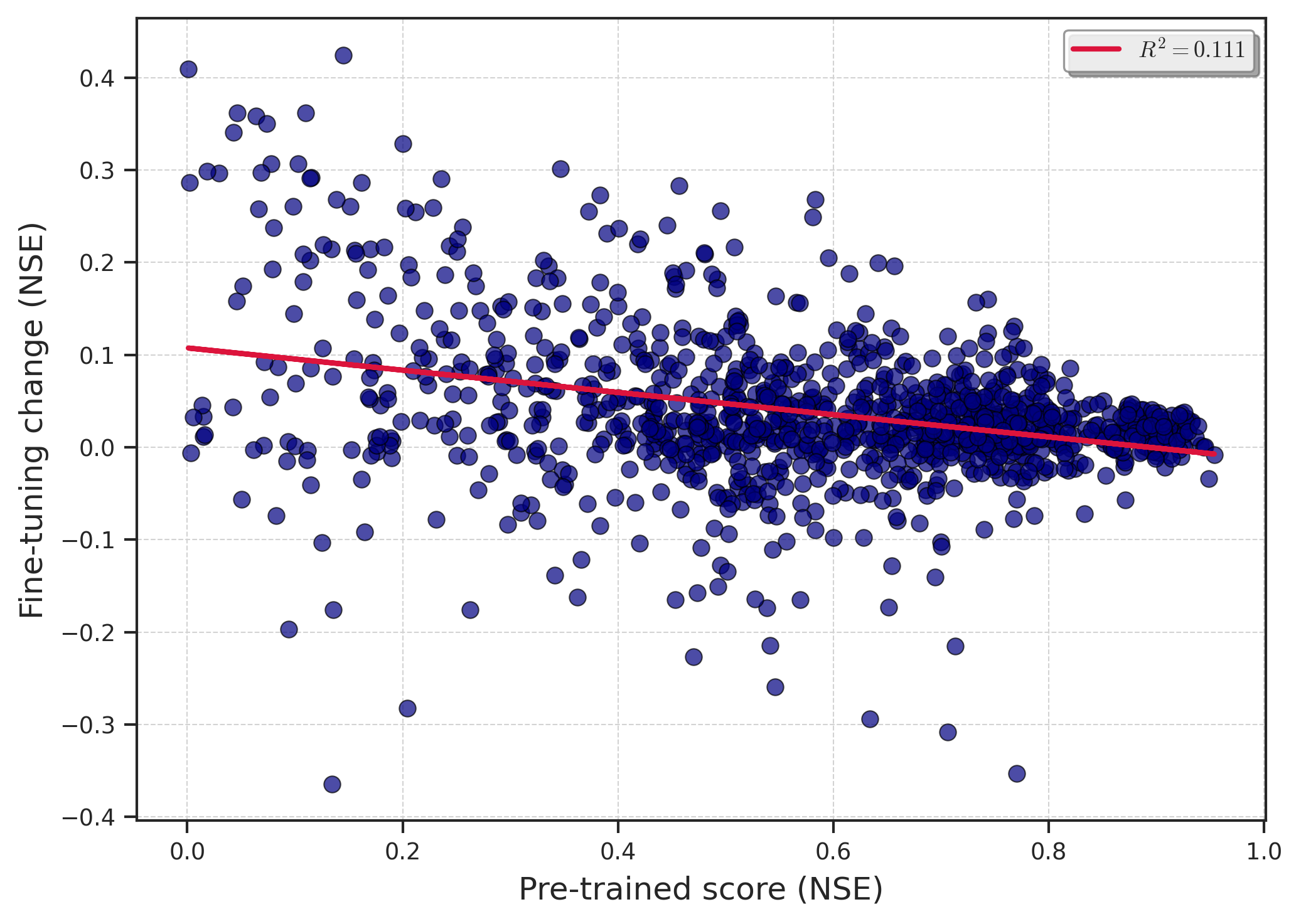}      \label{fig:skill-correlation}
    \end{subfigure}
    % Second subfigure
    \begin{subfigure}{0.49\linewidth}
        \centering
\includegraphics[width=\linewidth]{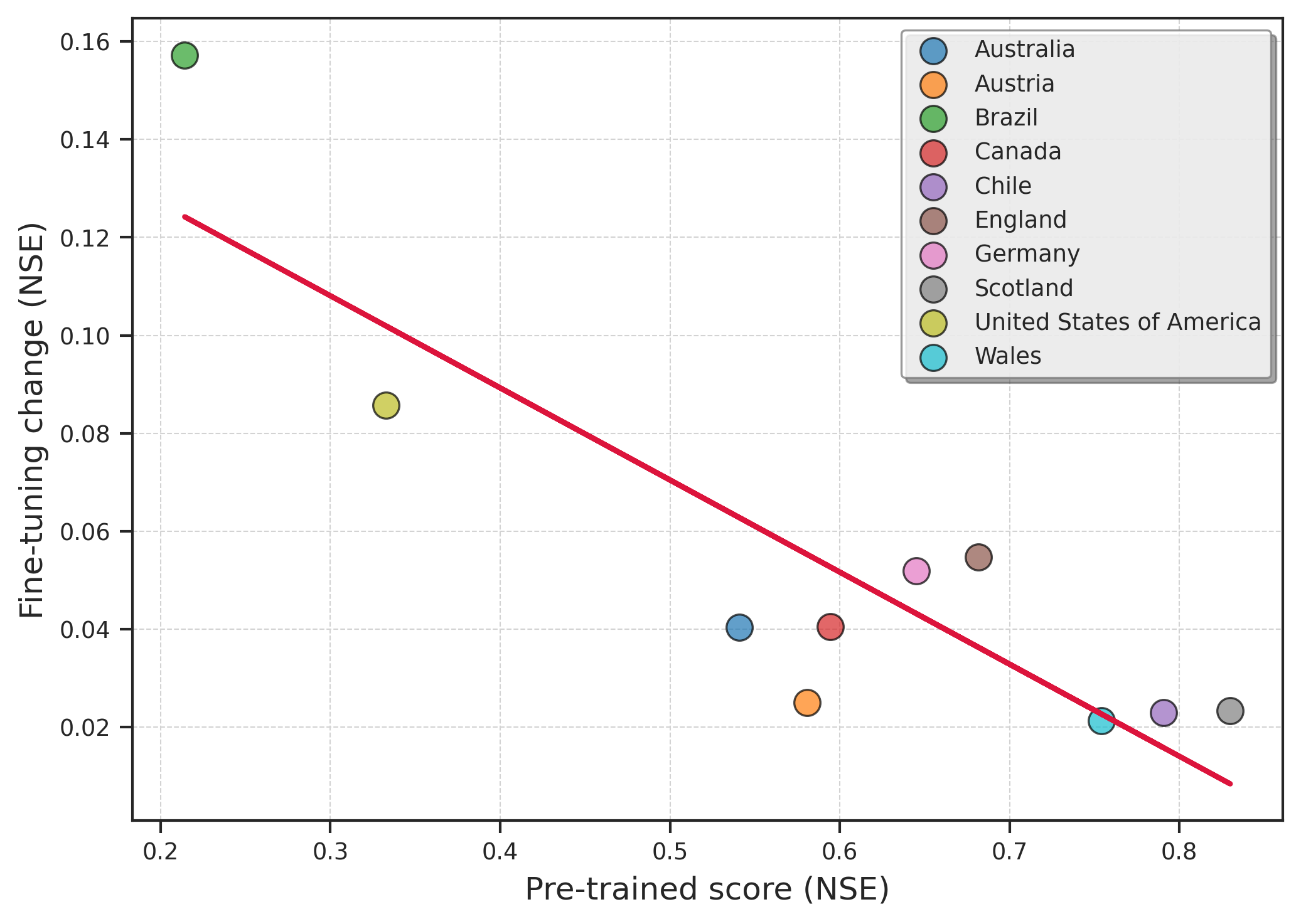}
\label{fig:country-correlation}
    \end{subfigure} 
\vspace{-15pt}
\caption{Correlation between fine-tuning improvements and pre-trained model skill across all basins (left) and aggregated across all countries (right). There is a negative correlation between pre-trained prediction skill and the improvement attained from fine-tuning.}
\label{fig:combined-correlation}
\end{figure}

Apart from just knowing the overall skill improvements, it is desirable to know where or under what conditions fine-tuning is more or less valuable. We show the relationship between fine-tuning improvement and original prediction score from the pre-trained model in figure \ref{fig:combined-correlation}, which is negative. Hence, basins on which the pre-trained models has a lower prediction skill tend to benefit more from fine-tuning. The fraction of variance explained by this trend is 11.1\%.
\begin{comment}
    should this fraction of variance (R-value) be commented on?
\end{comment}
Figure \ref{fig:combined-correlation} also shows that improvement through fine-tuning is still noisy and not guaranteed -- for $22\%$ of (base model, basin) pairs, fine-tuning even impacts performance negatively. Looking at each basin individually (across all the models), the proportion of negative changes is lower at $14\%$. 

As also seen in figure \ref{fig:combined-correlation}, for every country in the dataset, there is a net positive improvement from fine-tuning. The figure also shows a stronger negative relationship between the pre-trained score and the fine-tuning improvements. In particular, for the pre-trained model's worst-performing countries (Brazil \& the US), fine-tuning improvements are substantially higher compared to the other countries. Future work will correlate fine-tuning skill changes with the geographical and hydrological characteristics of the watersheds.
\vspace{-10pt}

\section{Conclusion \& Discussion}\label{conclusion}
We showed that fine-tuning a globally pre-trained model on individual basins leads to significant skill improvements, particularly in basins where a pre-trained global model struggles. Our results confirm the hypothesis -- informed by traditional hydrology -- that we can get more out of local data even when following the current best-practice for training ML-based hydrology models (i.e., training on large, diverse datasets). Furthermore, we found that basins and countries for which the pre-trained models struggle tend to gain the most from fine-tuning. Hence, we believe our method will be particularly helpful for national forecasters in regions that are currently underserved by global models. We provide a clear step-by-step guide for how to set up fine-tuning for forecasters in Appendix \ref{fine-tuning guide}.

In deciding whether to fine-tune a model, a matter that will likely be of particular importance to national forecasting agencies and hydrologists is the relationship between fine-tuning improvement and hydrological basin characteristics (e.g. aridity, area). By training a regression model on this task, one could find such relationship and hopefully also predict with accuracy for what basins fine-tuning will work. We expect that both of these questions will be important for national forecasters, and are questions left for future work.

\bibliographystyle{plainnat}
\bibliography{main}

\newpage
\appendix

\section{Fine-tuning guide for forecasters}\label{fine-tuning guide}
Here is a step-by-step guide for national forecasters and other regional operators who wish to fine-tune global models on their own data. An even more detailed guide can be found in our repository \url{https://github.com/EmilRyd/Fine-Flood-Forecasts.git}.

\begin{enumerate}
    
        \item \textbf{Download Caravan:} Information on the Caravan dataset can be found on Github at \url{https://github.com/kratzert/} and the actual dataset is downloaded from Zenodo at \url{https://zenodo.org/records/14673536}
        \item \textbf{(Optional): Add your own data}: If the basin data you want to fine-tune on is already in Caravan, then you can skip this step. Otherwise, add the basins (together with streamflow labels) to the Caravan dataset. Please note that you do not need to add this to the public dataset, you can just format your data into your local copy of Caravan and run it there. A complete tutorial for formatting your data into Caravan is at \url{https://github.com/kratzert/Caravan/wiki/Extending-Caravan-with-new-basins}.
        \item \textbf{Fine-tune on your data:} Follow the instructions in our repository to fine-tune your pre-trained model on your chosen basins. Our code automatically downloads a global pre-trained model which we have open-sourced, and provides all the necessary code to fine-tune this model on any individual basins (either already in Caravan or that you add into the Caravan format).
        
    \end{enumerate}

\section{Training details}\label{training setup}
\subsection{Model configuration and pre-training setup}\label{model config}
We use the Neuralhydrology library \citep{kratzert2022neuralhydrology} to pre-train 8 instances of our base model. We follow very closely the model configuration used by Roi-Cohen and Morin \citep{roi_cohen_morin_2024}, changing only the batch size from 128 to 256 to speed up training.  Our base model is an LSTM with a single hidden layer of size 256, with a linear embedding layer of 10 neurons for the static catchment attributes, and a linear head (see the Neural Hydrology documentation for more details). We trained our model for 40 epochs, with a learning rate of $5 \times 10^{-5}$ for epochs $1-30$, and $5 \times 10^{-6}$ for epoch $31-40$. We used NSE as the loss.
The precise model configuration is in the model config file in our repository (after downloading the pre-trained model from Hugging Face) for full details.

\subsection{Fine-tuning setup}\label{fine-tuning scheme}
We randomly sampled 159 basins from our full dataset. For each basin, we then performed a hyperparameter sweep over a set of fine-tuning parameters using the Tree-Structured Parzen Estimator \citep{Watanabe2023TreestructuredPE} within the hyperopt library \citep{Bergstra2013}. We ran 50 evaluations for every basin. We followed the same procedure for our single-basin trained models, with some additional hyperparemeters to vary the model size to avoid overparametrizing our model. Our hyperparameter search space is:

\begin{table}[h]
    \centering
    \begin{tabular}{lcc}
        \toprule
        \textbf{Hyperparameter} & \textbf{Fine-tuning} & \textbf{Training (single basin)} \\
        \midrule
        Epochs & $[1, 40]$ & $[1, 40]$ \\
        Learning Rate (epoch 1-20) & $[10^{-5}, 10^{-3}]$ & $[10^{-5}, 10^{-3}]$ \\
        Learning Rate (epoch 21-40) & $[10^{-6}, 10^{-4}]$ & $[10^{-6}, 10^{-4}]$ \\
        Loss \footnotemark[2]& \{NSE, MSE, RMSE\} & \{NSE, MSE, RMSE\} \\
        Fine-tuning modules & \{Full, Head\} & N/A \\
        Hidden size & N/A & \{16, 32, 64\} \\
        \bottomrule
    \end{tabular}
    \caption{Hyperparameter search space for fine-tuning.}
    \label{tab:hyperparam_search}
\end{table}
\footnotetext[2]{Here, ``Loss" only changes the loss that is calculated during gradient descent (for fine-tuning), but the parameter sweep still always evaluates against the NSE value on the validation set to find the best set of fine-tuning hyperparameters. In our public repository, however, the loss is set to be the same on both the validation and training set.}

For ``Fine-tuning modules", ``Full" means that we fine-tune all the weights from the pre-trained model (except the initial embedding layer for the statics attributes), and ``Head" means that fine-tune only the head, which is just a linear layer. We fine-tune each of our 8 base models on the 159 basins, and then average over the fine-tuning values for each of the basins.

\section{More results}\label{more results}
Above we reported NSE and KGE scores of our various models. These metrics measure correlations (in some sense) between simulated and predicted hydrographs, but do not capture all important aspects of hydrological prediction. For a more complete picture, we report several other common performance metrics. For more details on these metrics, see Gauch et al \citep{Gauch2023}.

\begin{comment}
    To Grey: Please let me know who to cite for these metrics
\end{comment}

\begin{table}[h]
    \centering
    \begin{tabular}{ l l l l }
        \toprule
        \textbf{Metric} & \textbf{Single-basin} & \textbf{Pre-trained} & \textbf{Fine-tuned} \\
        \midrule
        NSE (mean) & $0.358 \pm 0.106$ & $0.473 \pm 0.019$ & \textbf{0.541} $\pm 0.014$ \\
        NSE (median) & $0.541 \pm 0.025$ &  $0.582 \pm 0.016$ & $\textbf{0.627} \pm 0.008$ \\
        MSE (mean) & $2.338 \pm 0.297$ & $2.227 \pm 0.020$ & $\textbf{2.016} \pm 0.098$ \\
        MSE (median) & $1.265 \pm 0.117$ & $1.080 \pm 0.076$ & $\textbf{0.963} \pm 0.021$ \\
        RMSE (mean) & $1.271 \pm 0.070$ & $1.220 \pm 0.006$ & $\textbf{1.153} \pm 0.023$ \\
        RMSE (median) & $1.125 \pm 0.068$ & $1.039 \pm 0.037$ & $\textbf{0.982} \pm 0.021$ \\
        KGE (mean) & $0.331 \pm 0.197$ & $0.520 \pm 0.007$ & $\textbf{0.599} \pm 0.026$ \\
        KGE (median) & $0.609 \pm 0.022$ & $0.683 \pm 0.010$ & $\textbf{0.711} \pm 0.007$ \\
        Pearson $r$ (mean) & $0.708 \pm 0.017$ & $0.758 \pm 0.002$ & $\textbf{0.773} \pm 0.005$ \\
        Pearson $r$ (median) & $0.740 \pm 0.015$ & $0.795 \pm 0.005$ & $\textbf{0.804} \pm 0.004$ \\
        $\alpha$-NSE (mean) & $0.728 \pm 0.021$ & $\textbf{0.881} \pm 0.009$ & $0.842 \pm 0.006$ \\
        $\alpha$-NSE (median) & $0.749 \pm 0.035$ & $\textbf{0.880} \pm 0.005$ & $0.868 \pm 0.013$ \\
        $\beta$-NSE (mean) & $0.053 \pm 0.027$ & $0.059 \pm 0.006$ & $\textbf{0.033} \pm 0.005$ \\
        $\beta$-NSE (median) & $\textbf{0.004} \pm 0.005$ & $0.029 \pm 0.003$ & $\textbf{0.004} \pm 0.001$ \\
        $\beta$-KGE (mean) & $1.228 \pm 0.199$ & $1.173 \pm 0.009$ & $\textbf{1.085} \pm 0.026$ \\
        $\beta$-KGE (median) & $\textbf{1.004} \pm 0.042$ & $1.041 \pm 0.004$ & $1.007 \pm 0.016$ \\
        Peak timing (mean) & $0.750 \pm 0.043$ & $0.647 \pm 0.012$ & $\textbf{0.609} \pm 0.011$ \\
        Peak timing (median) & $0.625 \pm 0.040$ & $0.537 \pm 0.009$ & $\textbf{0.500} \pm 0.012$ \\
        Missed-Peaks (mean) & $0.605 \pm 0.014$ & $0.559 \pm 0.005$ & $\textbf{0.548} \pm 0.004$ \\
        Missed-Peaks (median) & $0.591 \pm 0.020$ & $0.546 \pm 0.008$ & $\textbf{0.537} \pm 0.007$ \\
        Peak-MAPE (mean) & $52.10 \pm 1.40 $ & $46.71 \pm 0.60$ & $\textbf{45.54} \pm 0.48$ \\
        Peak-MAPE (median) & $50.40 \pm 1.77$ & $45.18 \pm 0.61$ & $\textbf{44.18} \pm 0.61$ \\
        \bottomrule
    \end{tabular}
    \caption{Performance metrics with mean ± std/$\sqrt{n}$ and median ± MAD/$\sqrt{n}$ (mean absolute deviation)}
    \label{tab:metrics}
\end{table}

\end{document}